\documentclass[letterpaper, 10 pt, conference]{ieeeconf} 

\IEEEoverridecommandlockouts

\usepackage{graphicx}
\usepackage{textgreek}
\usepackage{amsmath}
\usepackage{caption}
\usepackage{subcaption}
\usepackage{booktabs}
\usepackage{multirow}
\usepackage{array}
\usepackage[colorlinks=true,linkcolor=blue, urlcolor=black]{hyperref}
\usepackage{algpseudocode}
\usepackage{algorithm}

\usepackage{xcolor} 

\title{\LARGE \bf
SwarmGear: Heterogeneous Swarm of Drones with Reconfigurable Leader 
Drone and Virtual Impedance Links for Multi-Robot Inspection
}

\author{Zhanibek Darush, Mikhail Martynov, Aleksey Fedoseev, Aleksei Shcherbak, and Dzmitry Tsetserukou
 \thanks{*The reported study was funded by RFBR and CNRS according to the research project No. 21-58-15006.}
\thanks{The authors are with the Intelligent Space Robotics Laboratory, Space CREI, Skolkovo Institute of Science and Technology, Moscow, Russian Federation.
 {\tt \{zhanibek.darush, mikhail.martynov, aleksey.fedoseev, aleksei.shcherbak, d.tsetserukou\} @skoltech.ru}}
}

\begin{document}

\maketitle
\thispagestyle{empty}
\pagestyle{empty}


\begin{abstract}

The continuous monitoring by drone swarms remains a challenging problem due to the lack of power supply and the inability of drones to land on uneven surfaces. Heterogeneous swarms, including ground and aerial vehicles, can support longer inspections and carry a higher number of sensors on board. However, their capabilities are limited by the mobility of wheeled and legged robots in a cluttered environment. 

In this paper, we propose a novel concept for autonomous inspection that we call SwarmGear. SwarmGear utilizes a heterogeneous swarm that investigates the environment in a leader-follower formation. The leader drone is able to land on rough terrain and traverse it by four compliant robotic legs, possessing both the functionalities of an aerial and mobile robot. To preserve the formation of the swarm during its motion, virtual impedance links were developed between the leader and the follower drones.


We evaluated experimentally the accuracy of the hybrid leader drone's ground locomotion. By changing the step parameters, the optimal step configuration was found. Two types of gaits were evaluated. The experiments revealed low crosstrack error (mean of 2 cm and max of 4.8 cm) and the ability of the leader drone to move with a 190 mm step length and a 3 degree standard yaw deviation. Four types of drone formations were considered. The best formation was used for experiments with SwarmGear, and it showed low overall crosstrack error for the swarm (mean 7.9 cm for the type 1 gait and 5.1 cm for the type 2 gait).

The proposed system can potentially improve the performance of autonomous swarms in cluttered and unstructured environments by allowing all agents of the swarm to switch between aerial and ground formations to overcome various obstacles and perform missions over a large area.


\end{abstract}


\section{Introduction}
 
In recent years, the impact of drones continues to increase in a wide scope of monitoring and inspection tasks. High mobility allows drones to overcome dense obstacles and collect data from areas inaccessible to the unmanned ground vehicles (UGV).  
\begin{figure}[h!]
\centering
 \includegraphics[width=1.0\linewidth]{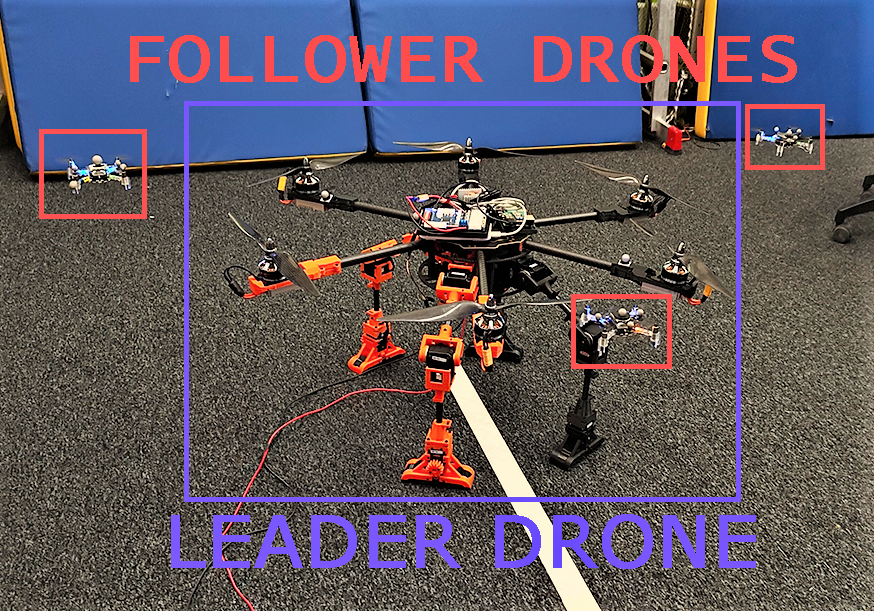}
 \caption{SwarmGear system during inspection.}
 \label{fig:main}
 \vspace{-0.2em}
\end{figure}
Intelligent swarms of unmanned aerial vehicles (UAVs) possess much higher capabilities for autonomous inspections, adapting to dynamic uncertain environments and complex tasks and collecting data from extensive areas \cite{Zhou_2020}. Many approaches of swarm formation control, path planning and data collection were proposed and developed for homogeneous swarms \cite{Abdelkader_2021}. However, the heterogeneous swarms of agents are shown to have better performance in missions than homogeneous swarms of similar size due their complimentary skills to take advantage of each configuration \cite{Scheutz_2005, Arnold_2020}.  
 
Several concepts of the heterogeneous swarms were proposed in previous research. For example, Arnold et al. \cite{Arnold_2019} suggested a new approach of object classification with swarm of UAVs carrying different sensors. Uryasheva et al. \cite{Uryasheva_2019} developed a task dispatching system based on greedy algorithm for heterogeneous UAV painting in digital arts.
Aside from heterogeneous swarms of drones, several papers proposed swarms with cooperation of multiple drones and mobile robots. 
Kojima et al. \cite{Kojima_2022} proposed a swarm behavior approach for  industrial plant inspection with robotic rovers and UAVs.
However, the applications of multi-agent heterogeneous swarms are limited by the mobility of the ground robots in cluttered environments and their ability to overcome uneven terrain. Few project tend to address this problem, e.g., cooperative object search algorithm by Salas et al. \cite{Salas_2021}, where drones are utilized to help mobile robot in path planning. Another approach was introduced by Pushp et al. \cite{Pushp_2022}, where UAVs are delivering mini-UGVs for exploration in cluttered environments.

In this paper we propose a novel multi-agent system SwarmGear (Fig. \ref{fig:main}), utilizing the capability of the hybrid drone to avoid both ground and aerial obstacles to increase the scope of the swarm applications.

\section{Related Works}
The most important feature of the SwarmGear is the ability of the drone to switch between tasks on the ground and in the air. The application of the robot is expanding with the possibility of ground locomotion. When developing the platform, we considered the work done by other groups of researchers exploring UAVs equipped with robotic arms and chassis.


Many researchers have explored aerial manipulation with objects.
Chen et al. \cite{Chen_2022} developed a drone equipped with robotic arm for aerial manipulation based on multi-objective optimization. Aerial manipulation through VR interface and drone equipped with 4-DoF robotic arm was proposed by Yashin et al. \cite{Yashin_2019}.


In addition to using robotic arms for manipulation tasks, its were also used by researchers as a locomotion system. Cizek et al. \cite{Cizek_2018} presented hexapod robot and Hooks et al. \cite{Hooks_2018} developed four-legged platform ALPHRED which could be integrated with UAV.



Pedipulators can be used as landing gear for a safe landing on an uneven surface. This idea has been implemented by DroneGear hybrid drone developed by Sarkisov et al. \cite{Sarkisov_2018}.


Industrial inspection typically includes a wide scope of operations needed to be performed on a large unstructured area. Therefore, the ability of a swarm to navigate while maintaining strict formation is highly required. Several concepts of formation control and swarm behavior were proposed prior to this work  \cite{Handayani_2015}. For example, the potential field method for heterogeneous swarm formation and path planning was introduced by Barnes et al. \cite{Barnes_2006}. Later, the decentralized coalition formation was developed by Sujit et al. \cite{Sujit_2015} for swarms with limited communication range. Another approach considering the limited field of view of the drones in swarm was proposed by Yang et al. \cite{Yang_2018} with V-shaped formation structure. Independent swarm intelligence solution was developed by Chen et al. \cite{Chen_2022_F} for the missile swarms. Leader-follower formation with artificial potential fields (APF) was proposed by Gupta et al. \cite{Gupta_2022} for the visually-localized swarm landing on the moving platform. On the other hand, Gao et al. \cite{Gao_2022} suggested the fully distributed control strategy based on minimal virtual leader formation.

The SwarmGear heterogeneous system, inspired by the previous concept of the DroneGear, proposes a new walking platform with 12 DoF and higher accuracy of locomotion for the leader drone. Unlike prior projects, we investigate not only the capability of the hybrid drone motion, but its joint mission with a swarm of the following drones. This The multi-agent capability of the SwarmGear can potentially increase the area of autonomous exploration and allow hybrid heterogeneous teams to carry different sensors for target detection.

\section{System Overview}

\begin{figure}[htb!]
\centering
 \includegraphics[width=1.0\linewidth]{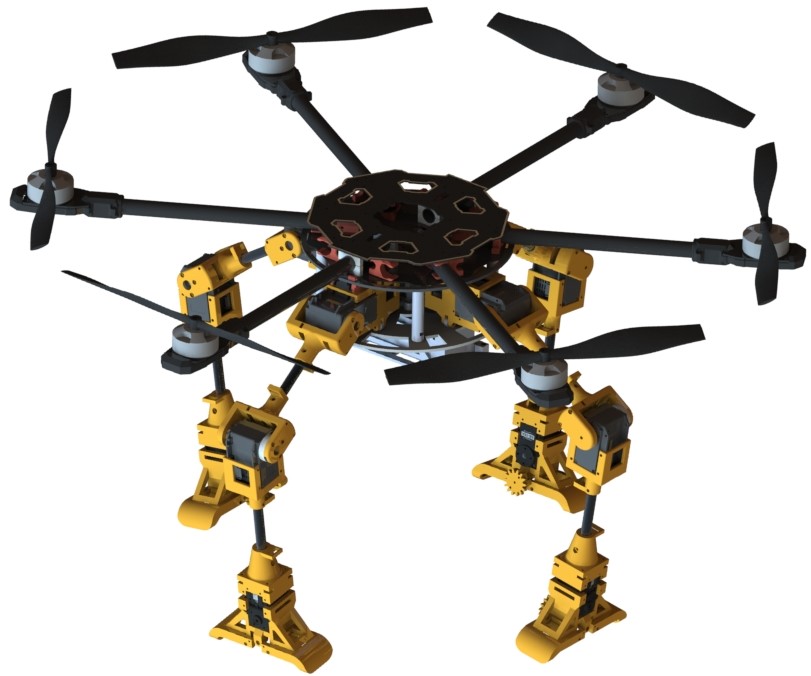}
 \caption{Render of the SwarmGear leader drone.}
 \label{fig:sover}
 \vspace{-0.2em}
\end{figure}

\subsection{Mechanics of leader drone}
The landing gear consists of four arms with three DoFs, which are at right angles to the central axis of the robot. The legs are moved by Dynamixel MX-106 and Dynamixel MX-28 servomotors in the hip joints and by Dynamixel MX-64 servomotors in the arm joints. In addition, the arm joints contain built-in optical torque sensors which consist of compression springs and an encoder located in a plastic housing. Each leg has a passive footrest which is connected to the tip of the forearm link. All parts of the mounts are made of PLA (polylactide) light weight. 
While the base consists of 2 carbon disks  3 mm thick caring the legs and 6 axises are attached (Fig.~\ref{fig:sover}). Expanded characteristics of the drone are shown in Table 1.

\subsection{Electronics and Software of the Leader Drone}
\begin{figure}[htb!]
\centering
 \includegraphics[width=1.0\linewidth]{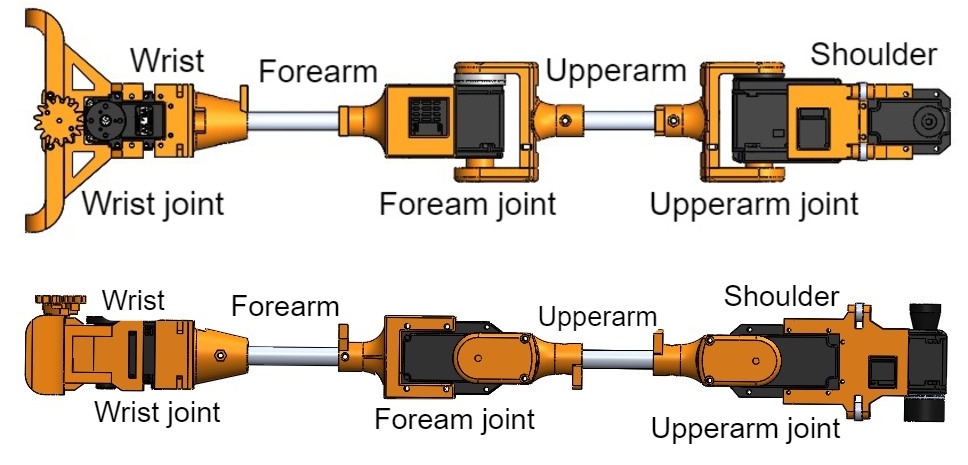}
 \caption{CAD design of the leader drone’s pedipulator.}
 \label{fig:leg}
 \vspace{-0.2em}
\end{figure}

The system consists of three parts: Unity as simulation environment, STM32 as microcontroller and Raspberry Pi 4 as command receiver by Robot Operating System (ROS). Unity provides simulation of robot motion control with visualization. Commands from Unity are received by ROS, which sends them to STM32. STM32 can read data from the encoders and control Dynamixel servomotors. STM32 returns feedback to Unity with actual information about current parameters of servomotors (voltage, rotation angle, and temperature). 
All calculations of the robot's movement are provided by C\# scripts in the Unity game engine. Then data of the angular position of servomotors are sent as commands. The CAD model of the landing gear is shown in Fig.~\ref{fig:leg}. 

\begin{table}[h!]
\centering{
\caption{Technical characteristics of SwamGear}
\setlength{\tabcolsep}{10pt} 
\renewcommand{\arraystretch}{1}
\begin{tabular}{ | c | c |  }
\hline
\label{Table Result 1}
\begin{tabular}{p{7cm}cp{3cm}} \end{tabular} 
Maximum load of T-motor MN4010& 2.2 kg   \\\hline
Maximum torque of Dynamixel MX106 at 12V & 8.4 N m  \\\hline
Maximum torque of Dynamixel MX64 at 12V  & 6.0 N m  \\\hline
Maximum torque of Dynamixel mx28 at 12V  & 2.5 N m  \\\hline
Maximum torque of Dynamixel ax12 at 12V  & 1.5 N m  \\\hline
Weight of one leg  & 0.58 kg  \\\hline
Length of Shoulder & 0.093 m \\\hline
Length of Upperarm & 0.154 m  \\\hline
Length of Forearm  & 0.206 m  \\\hline
Length of Wrist    & 0.044 m  \\\hline
Battery capacity   & 8000 mAh,22.2V  \\\hline
\end{tabular}}
\end{table}

\subsection{Kinematics of the leader drone locomotion}
The leaded drone has two types of gait. They are based on various algorithms and motion logic. With the first type of walking, taking into account the initial parameters, the trajectory of the limbs is pre-calculated, and the limbs move along a closed trajectory a given number of times. For the second type of gait, the algorithm depends on the initial parameters of the position of the limbs and their changes during movement.
\subsubsection{Walking Gait the first type}
In this type of gait, the limbs move along a closed trajectory synchronously. On the ground, the limbs move in a straight line, however, when moving the limb to the beginning of a straight trajectory (the upper part of the trajectory), the limbs move along part of the Archimedean spiral.

The robot's design is symmetrical with respect to the center of the base . Thus, the trajectories for the two limbs in front are the same, as well as for the two rear ones, with the only difference that the trajectories of the hind limbs are reflected relative to the center of the base. In addition, the front two limbs move synchronously clockwise, but while one of them is a 12 o'clock on trajectory, the second one follows a 6 o'clock on trajectory. The hind limbs move the same, thus, two opposite limbs are on the ground and move the robot forward, passing along the straight part of the trajectory, while the other two limbs move forward along the upper part of the trajectory, which then continues the robot's movement so that the other two limbs can return to the beginning of the trajectory.
When developing an algorithm for robot locomotion, it is necessary to know the height of the robot, which depends on the initial angle of upperarm joint, in order to calculate the length of the straight part of the trajectory, the ends of which are closed using the Archimedes spiral



Robot height was obtained by the following equation:
\begin{equation}
 \label{robot height}
 H = \sin\beta_{init} \cdot l_{UA} + l_{FA}
\end{equation}
where $l_{FA}$ is the length of the forearm, $l_{UA}$ is the length of the upperarm and $\beta_{init}$ is the initial standing angle of upperarm joint.
To find the extreme point of the straight part of the trajectory, the position of the end effector for the straightened limb was calculated.
To prevent the robot from falling, the end of the straight part of the trajectory is considered to be the projection of the attachment point to the base on the ground.

The extreme point of the straight part of the trajectory was obtained by the following equation:
\begin{equation}
 \label{segment beginning}
    x_0 = \sqrt{(l_{UA}+l_{FA})^2-H^2}
\end{equation}
The projection of the attachment point of the limb to the base considered as zero point.

Inverse kinematics of the robot limbs was calculated as follows: 
\begin{multline}
 \label{Inverse Kinematics}
 \cos(\gamma) = (x^2+y^2- l_{UA}^2-l_{FA}^2)/(2l_{UA}l_{FA})\\
 \sin(\gamma) = \sqrt{(1-\cos(\gamma)^2)}\\
 \gamma = -\arccos(\cos(\gamma))\\
 \beta = arctan2(y,x) + \\ + arctan2(l_{FA}\sin(\gamma),l_{UA}+l_{FA}\cos(\gamma))
\end{multline}

\subsubsection{Walking Gait the second type}
This type of gait is based on the fact that the robot relies on two limbs when walking and pushes itself forward by turning the shoulder joint. First, the robot moves its two opposite limbs forward and places them vertically on the ground. Next, the robot begins to lift itself with the help of the forearm and upperarm joints and simultaneously rotates the shoulder joint, thereby lifting and moving itself forward.

It is assumed that the limbs at the fulcrum are stationary and can only rotate around the vertical axis. Then, in order to physically propel itself forward, the robot needs to lift itself up to compensate for the platform shift that occurs when the limb rotates around the fulcrum. Since two opposite limbs rotate, this shift is directed perpendicular to the direction of movement, which can be compensated by reducing the length of the limb along the horizontal axis, i.e. changing the angles in the forearm and upperarm joints.
Thus, in order to avoid compression of the robot's base, the robot compensates for the horizontal shift (Eq.~\ref{shift_horizontal}), which can be calculated as follows:


\begin{equation}
 \label{shift_horizontal}
  \delta = 2(1-\cos\alpha_{sh})l_p
\end{equation}
where $\alpha_{sh}$ is the supporting limb's rotation angle of the shoulder joint which increasing during first half of step and then decreasing, $l_p$ is the projection of limb length on the ground.
Thus, the robot lifts itself up, then lowers itself to its original position.

Robot compensates the compression of the base as follows (\ref{shift_vertical}):
\begin{equation}
 \label{shift_vertical}
  \delta = (\cos \beta_{init} - \cos (\beta_{init}+\xi)) \cdot l_{UA}
\end{equation}
where $\beta_{init}$ is the initial standing angle equals $30^{\circ}$ for upper arm joint and $60^{\circ}$ for forearm joint, $\xi$ is the compensating angle for these joints, $l_{UA}$ is the length of upper arm.
$l_{UA}$ refers to the projection $l_{p}$ as $l_{UA}\cos \beta_{init}$. 

From equations (\ref{shift_horizontal}),(\ref{shift_vertical}), we obtained the relationship between the angle of rotation of the shoulder joint and the compensating angle:
\begin{equation}
 \label{comparizon equation}
  2(1-\cos\alpha)l_p = (\cos \beta_{init} - \cos (\beta_{init}+\xi)) l_{UA}
\end{equation}

Thus, by expressing $\xi$ from the equation (\ref{comparizon equation}), we were able to obtain forward motion.
\begin{equation}
 \label{xi}
  \xi = \arccos((2\cos \alpha_{sh} - 1)  \cos \beta_{init}) -\beta_{init}
\end{equation}

The limbs located in front and behind are raised at the moment of the step.
\begin{figure}[!t]
\centering
\includegraphics[width=3.5in]{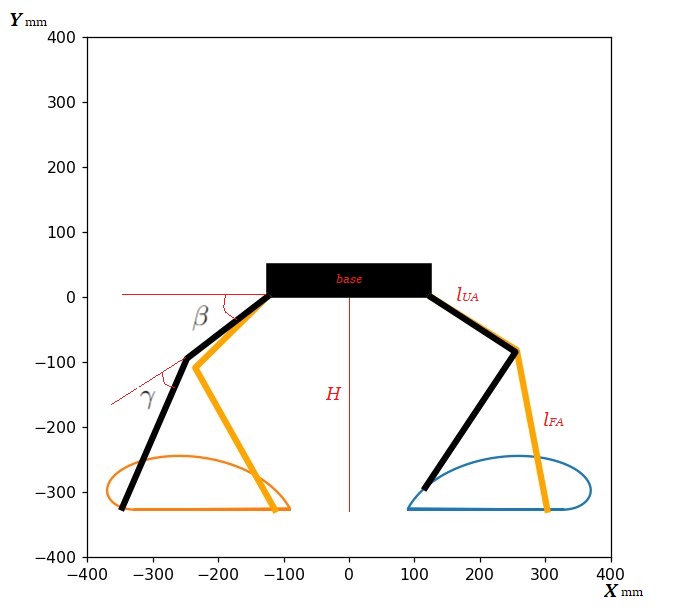}
\caption{Limbs motion trajectories for the first type of gait.}
\label{fig_crab}
\end{figure}

\subsection{Impedance-based formation control of the follower drones}

\begin{figure}[htb!]
\centering
 \includegraphics[width=1.0\linewidth]{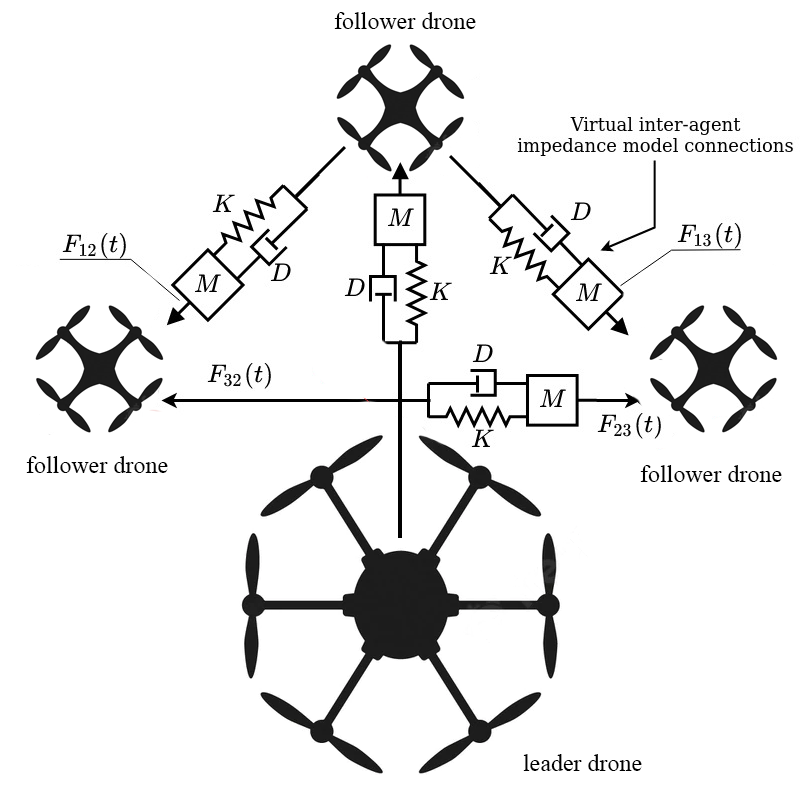}
 \caption{The topology and parameters of impedance links to achieve safe flight and compliant interaction.}
 \label{fig:impedance}
 \vspace{-0.2em}
\end{figure}

The position and velocity of each drone in the swarm in leader-follower formation is naturally dependant on the state of the leader drone. In order to achieve a sufficient level of flight accuracy and swift response of the follower drone swarm, we applied the impedance control model utilized by Tsykunov et al. \cite{Tsykunov_2019}. The virtual mass-spring-damper model that defines follower drone position correction based on leader drone and potential obstacles is shown in Fig. \ref{fig:impedance}. 

The configuration of these connections is determined by the specific application and desired swarm formation. The impedance provided by the virtual damped spring model prevents collisions within the swarm agents and with the user, thus providing a safe environment for object interaction.

Each virtual mass-spring-damper link dynamics is calculated with the position-based impedance control approach introduced in \cite{Hogan_1984} and can be represented by the second-order differential equation:

\vspace{-0.4em}

\begin{equation}
 \label{eq:1}
 M\Delta{\ddot x} + D\Delta{\dot x} + K\Delta x = F_\text{ext}(t) 
\end{equation}

\noindent
where $M$ is the virtual mass, $D$ is the damping coefficient of the virtual damper, $K$ is the stiffness of the virtual spring, $\Delta x$ is the difference between the current drone position $x_c$ and the desired position $x_d$, and $F_\text{ext}(t)$ is the externally applied force. For human-drone connections, the force is calculated using a human state parameter. In our implementation of the model, the external force for human-drone connections $F_\text{human}(t)$ is calculated as directly proportional to the leader drone velocity $v_\text{leader}$ in order to ensure a smooth trajectory for the drones with proper orientations and positions following the hybrid drone, as defined in: 

\vspace{-0.4em}

\begin{equation}
 \label{eq:2}
 F_\text{human}(t) = K_v v_\text{leader}(t)
\end{equation}

\vspace{0.5em}

\noindent
where $K_v$ is the scaling coefficient, which can be selected to produce desirable feedback from the drones in response to the leader drone displacement.
In order to solve the second-order differential equation (\ref{eq:1}), we utilize the solution discussed in \cite{Tsykunov_2019}, which rewrites the impedance equation as a state-space representation for discrete-time as follows:

\begin{equation}
 \label{eq:3}
 \begin{bmatrix}
 \Delta \dot x \\
 \Delta \ddot x
 \end{bmatrix} =
 A \begin{bmatrix}
 \Delta x \\
 \Delta \dot x
 \end{bmatrix} +
 BF_\text{ext}(t)
\end{equation}

\noindent
where, $A = \begin{bmatrix}
0 & 1 \\
-\frac{K}{M} & -\frac{D}{M}
\end{bmatrix}$ and $B = \begin{bmatrix}
0 \\
\frac{1}{M}
\end{bmatrix}$. The model is further simplified by integrating this equation in a discrete time-space, which is given by:

\vspace{-0.2em}

\begin{equation}
 \label{eq:4}
 \begin{bmatrix}
 \Delta x_{k+1} \\
 \Delta{\dot x}_{k+1}
 \end{bmatrix} =
 A_d \begin{bmatrix}
 \Delta x_k \\
 \Delta{\dot x}_{\text{k, leader}}
 \end{bmatrix} +
 B_d F_\text{ext}^k
\end{equation}

\vspace{0.5em}

\noindent
where $A_d$ and $B_d$ are defined using the matrix exponential and found using the Cayley-Hamilton theorem. The required calculations are defined in:

\begin{equation}
 \label{eq:5}
 A_d = e^{\lambda T} \begin{bmatrix}
 (1 - \lambda T) & T \\
 -bT & (1 - \lambda T - aT)
 \end{bmatrix}
\end{equation}

\vspace{0.7em}

\begin{equation}
 \label{eq:6}
 B_d = -\frac{c}{d} \begin{bmatrix}
 e^{\lambda T}(1 - \lambda T) - 1 \\
 -bTe^{\lambda T}
 \end{bmatrix}
\end{equation}

\vspace{0.5em}

\noindent
where $a = -\frac{D}{M}$, $b = -\frac{K}{M}$, $c = \frac{1}{M}$, and $\lambda$ is the eigenvalue of the matrix $A$. By selecting the parameters of the impedance model ($M$, $D$, $K$, and $K_v$) in such a way that the model is critically damped, the model is further simplified by ensuring both that $A$ only has one eigenvalue, $\lambda = \lambda_1 = \lambda_2$, and that solution is real. This eigenvalue can then be found as the root of the characteristic equation of the matrix $A$, given in:

\vspace{-0.4em}

\begin{equation}
 \label{eq:7}
 \lambda^2 + 2\zeta \omega_n \lambda + \omega_n^2 = 0
\end{equation}

\vspace{0.5em}

\noindent
where $\omega_n = \sqrt{\frac{K}{M}}$, $\zeta = \frac{D}{2\sqrt{MK}}$. By finding the total applied force $F_\text{ext}(t)$ for a given virtual link the target position and velocity of each drone can be calculated using Eq. (\ref{eq:4}). Each drone in the swarm then follows a certain position with a given offset from the leader drone while avoiding collisions between the follower drones in the swarm.

\begin{figure}[htb!]
 \includegraphics[width=1.0\linewidth]{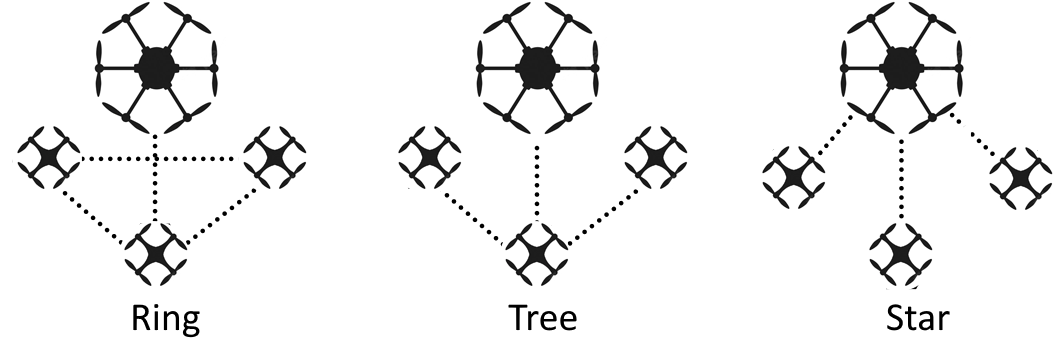}
 \caption{Evaluated impedance link topologies of drones in swarm for the impedance control model.}
 \label{fig:topology}
 \end{figure}

Both the impedance model parameters and the interlink topology configuration require additional investigation to ensure that the drone swarm can maintain a desirable formation. Impedance model parameters were calculated to satisfy a critically damped response (i.e. satisfying $\zeta = 1$), in addition, the APF method was tested. Three interlink network topologies were tested (Fig. \ref{fig:topology}).

Several simulations were run to observe both the positions and velocities of the drones. These tests demonstrated that the most desirable set of parameters for the SwarmGear formation was $M=1.9$, $D=12.6$, $K=20.88$, corresponding to a natural frequency $\omega_n = 3.3$.

\section{ Experimental evaluation}
\subsection{Evaluation of different impedance formations}
\begin{figure}
 \includegraphics[width=1.0\linewidth]{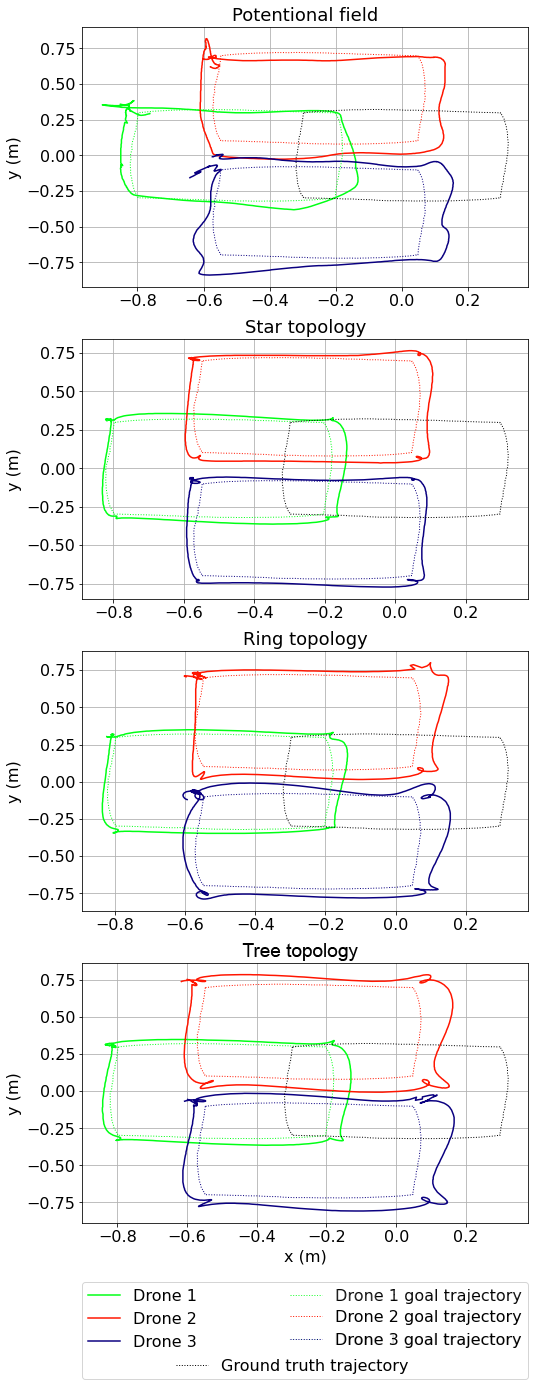}
 \caption{Trajectories of drones with impedance link topologies of the three-agent swarm with following parameters: M=1.9, D=12.6, K=20.88.}
 \label{fig:position_plots}
 \end{figure}

To test the real swarm behavior, we conducted an experiment with the drones following the UR3 robotic arm that performed a square trajectory. The experiment was conducted 3 times for each formation, after which an average result was calculated. Fig. \ref{fig:position_plots} shows the trajectories of the swarm following the robot's end-effector.

The experimental results showed that the artificial potential field method is the less accurate compared to the impedance control. The quantitative assessment is shown in Table \ref{tap:error}.

\begin{table}[h!]
\centering{
\caption{Positional error of drones with Impedance and Potential field swarm control algorithms.}
\setlength{\tabcolsep}{6pt} 
\renewcommand{\arraystretch}{1}
\begin{tabular}{ | c | c | c | c | c | }
\hline
\label{tap:error}
\begin{tabular}{p{1.5cm}cp{4cm}} Swarm control approach \end{tabular} & \multicolumn{2}{|c|}{Mean error, m} & \multicolumn{2}{|c|}{Max error, m} \\ \cline{2-3} \cline{4-5}
 & x & y & x & y \\ \hline
Impedance, Ring t.& 0.13 & 0.14 & 0.35 & 0.36 \\\hline
Impedance, Tree t. & 0.14 & 0.16 & 0.35 & 0.37 \\\hline
Impedance, Star t. & 0.10 & 0.11 & 0.27 & 0.29 \\\hline
Potential f. & 0.22 & 0.22 & 0.49 & 0.45 \\ 
\hline
\end{tabular}}
\end{table}

From these results, the MSE for the position of drones (using the difference in x and y coordinates from the expected trajectory) was found to be $0.10$ m for x and $0.11$ m for y for the Star topology, which is the best configuration tested. 

\begin{figure}[htb!]
 \includegraphics[width=1.0\linewidth]{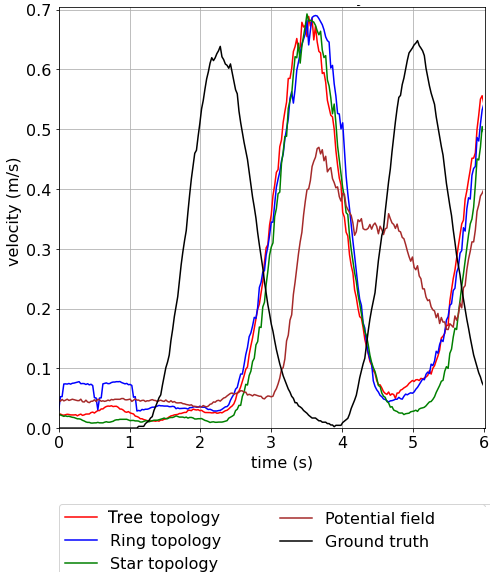}
 \caption{Mean velocities of the swarm drones in XY horizontal plane during testing of each topology.}
 \label{fig:velocity_plot}
 \vspace{-0.4em}
\end{figure}

The velocities of drones following a trajectory with the potential field method was significantly reduced to increase the stability of the drone flight (Table \ref{tap:speed}). The parameters of the Impedance Control method correspond to the speed of the robot. In Fig. \ref{fig:velocity_plot}, we observe a 1.3-second time delay for the impedance control.
 
The potential field method generates a more continuous motion with a lower acceleration value while significantly increasing the positioning error. Thus, for the experiments, we have chosen the impedance control behavior during inspection and the  artificial potential fields for approaching the initial position after swarm deployment. The smoother movement explains the latter choice towards a distant target.

\begin{table}[]
\centering{
\caption{Velocity of drones with Impedance and Potential field swarm control algorithms.}
\setlength{\tabcolsep}{6pt} 
\renewcommand{\arraystretch}{1}

\begin{tabular}{ | c | c | c | }
\hline
\label{tap:speed}
 Swarm control approach & Max velocity, m/s & Mean velocity, m/s\\ \hline
Impedance, Ring t.& 0.69 & 0.24  \\\hline
Impedance, Tree t. & 0.70 & 0.24 \\\hline
Impedance, Star t. & 0.69 & 0.22 \\\hline
Potential f. & 0.47 & 0.20  \\ \hline
Ground truth & 0.65 & 0.18  \\ \hline
\end{tabular}}
\end{table}

\subsection{Linear trajectory following by leader drone}
A series of experiments were conducted to find the optimal starting position of the robot for the Type 2 gait. Afterwards the accuracy was confirmed when moving in a straight line.
Various angular velocities of servomotors when walking are also considered. In addition, commands were also sent with different frequencies to find the most uniform movement.

\begin{figure}[htb!]
 \includegraphics[width=1.0\linewidth]{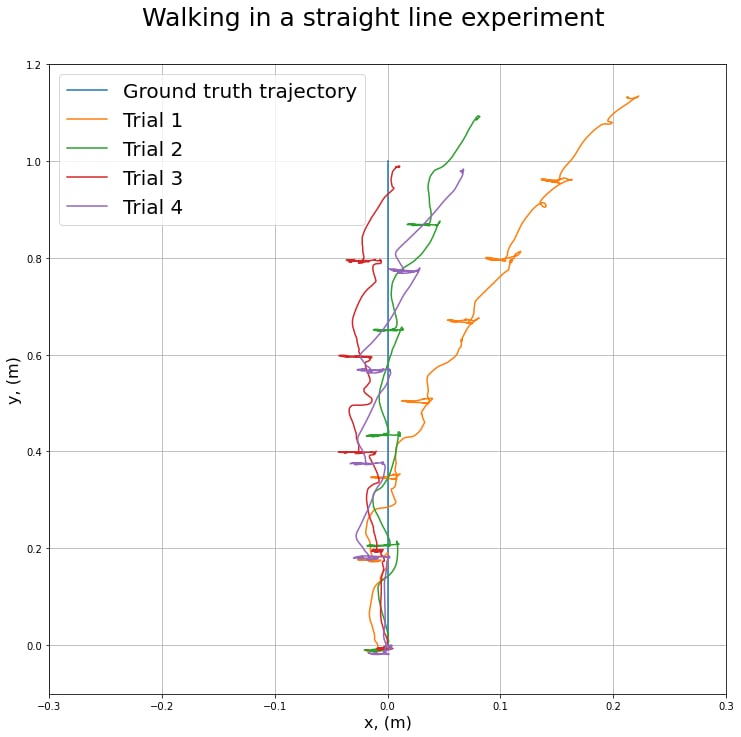}
 \caption{Linear trajectory following by the leader drone moving with Type 2 gait and $\beta = 30^{\circ}$.}
 \label{fig:results gamma 30}
 \vspace{-0.4em}
\end{figure}

\begin{figure}[htb!]
 \includegraphics[width=1.0\linewidth]{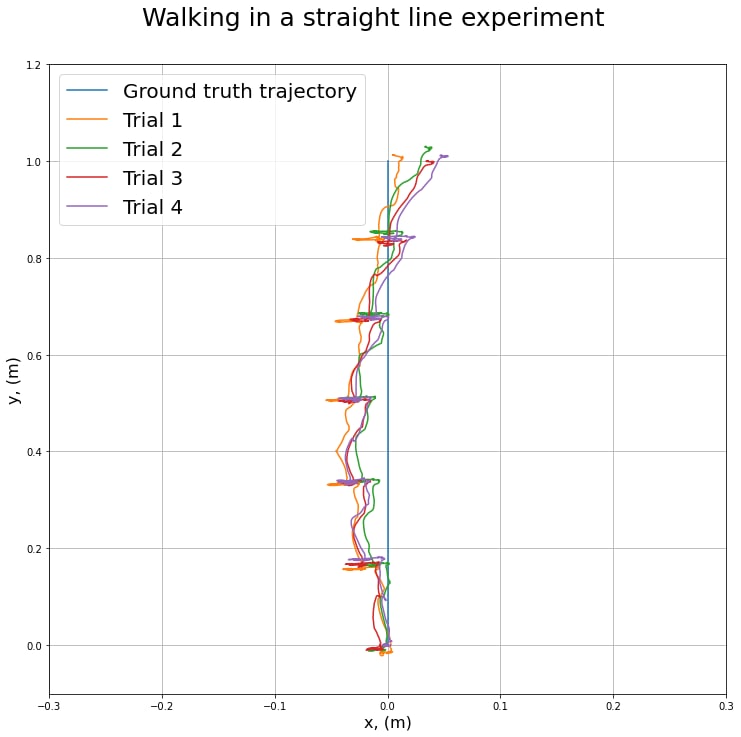}
 \caption{Linear trajectory following by the leader drone moving with Type 2 gait and $\beta = 45^{\circ}$.}
 \label{fig:results gamma 45}
 \vspace{-0.4em}
\end{figure}

The experiment consisted of a series of trials where the robot had to walk a meter forward in a straight line. The trajectory of movement can be seen in the figures \ref{fig:results gamma 30} and \ref{fig:results gamma 45}. Based on the results of the experiments the deviation from the trajectory was analyzed. The results are shown in the Table \ref{Table Result}.

\begin{table}[h!]
\centering{
\caption{Experiments setup and results.}

\setlength{\tabcolsep}{6pt} 
\renewcommand{\arraystretch}{1}
\begin{tabular}{ | c | c | c | c | c | }
\hline
\label{Table Result}
\begin{tabular}{p{1.5cm}cp{4cm}}  \end{tabular} & \multicolumn{2}{|c|}{Mean error, m} & \multicolumn{2}{|c|}{Max error, m} \\ \cline{2-3} \cline{4-5}
 & $\beta = 30^{\circ} $ & $\beta = 45^{\circ} $ & $\beta = 30^{\circ} $ & $\beta = 45^{\circ} $ \\ \hline
Trial 1 & 0.057 & 0.026 & 0.223 & 0.055 \\\hline
Trial 2 & 0.019 & 0.018 & 0.082 & 0.040 \\\hline
Trial 3 & 0.014 & 0.021 & 0.043 & 0.044 \\\hline
Trial 4 & 0.019 & 0.018 & 0.067 & 0.054 \\\hline
Overall & 0.027 & 0.021 & 0.104 & 0.048 \\
\hline
\end{tabular}}
\end{table}

The best result was achieved with the initial position of the upper arm at 45 degrees (q.v. $\beta$ Eq.~\ref{xi}), the angular speed of the servos at 45 degrees per second, the frequency of sending commands at 0.025 seconds. It was confirmed by a series of experiments that the deviation at the end trajectory point did not exceed 4 centimeters per meter.

\subsection{Linear trajectory following by swarm of drones with a Star impedance formation}
This experiment was conducted in order to see the deviation from the trajectory that occurs in the formation of drones when passing one meter of the path.

Two experiments were conducted for different types of gait. The robot's gaits differ in that with the first type of gait, the platform swings more, but the gait itself is continuous and smoother. The second type of gait is more accurate, but after each step the robot stops waiting for the command to take the next step.

Thus, we looked at how two different types of walking affect the formation of a swarm of drones and how much the sudden acceleration and stopping or rocking of the platform affects impedance-based formation control of the follower drones.

Based on the results of previous experiments Star topology have shown the best results so this this impedance control model was chosen for the experiment.

 \begin{figure}[htb!]
 \includegraphics[width=1.0\linewidth]{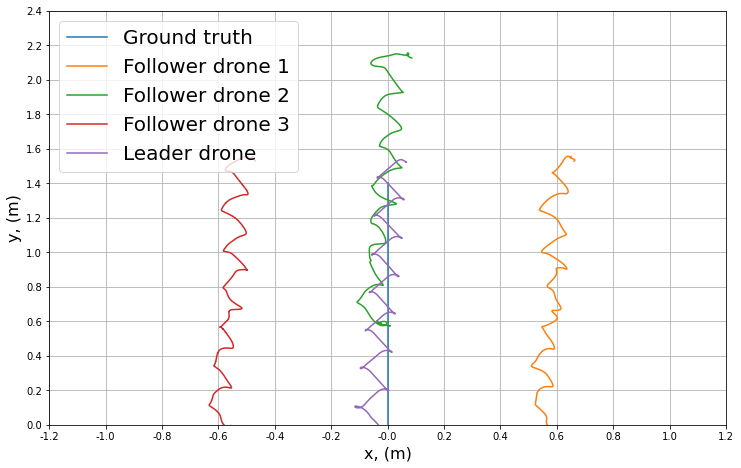}
 \caption{Linear trajectory following by a swarm, in which the leader drone is moving with Type 1 gait.}
 \label{fig:results dog walk}
 \vspace{-0.4em}
\end{figure}

Table \ref{Table Result 1} shows the RMSE of 3.3 cm and maximal error does of 12.9 cm after one meter of linear path following. 

\begin{table}[h!]
\centering{
\caption{Positional error of the swarm. Gait Type 1. Star formation}
\setlength{\tabcolsep}{6pt} 
\renewcommand{\arraystretch}{1}
\begin{tabular}{ | c | c | c | }
\hline
\label{Table Result 1}
\begin{tabular}{p{1.5cm}cp{4cm}} \end{tabular} & {RMSE, m} & {Max Error, m} \\\hline
Leader Drone & 0.033 & 0.090 \\\hline
Follower drone 1 & 0.029 & 0.066 \\\hline
Follower drone 2 & 0.049 & 0.112 \\\hline
Follower drone 3 & 0.059 & 0.129 \\\hline
Followers overall & 0.046 & 0.129 \\\hline
Overall & 0.039 & 0.129 \\\hline
\end{tabular}}
\end{table}

 \begin{figure}[htb!]
 \includegraphics[width=1.0\linewidth]{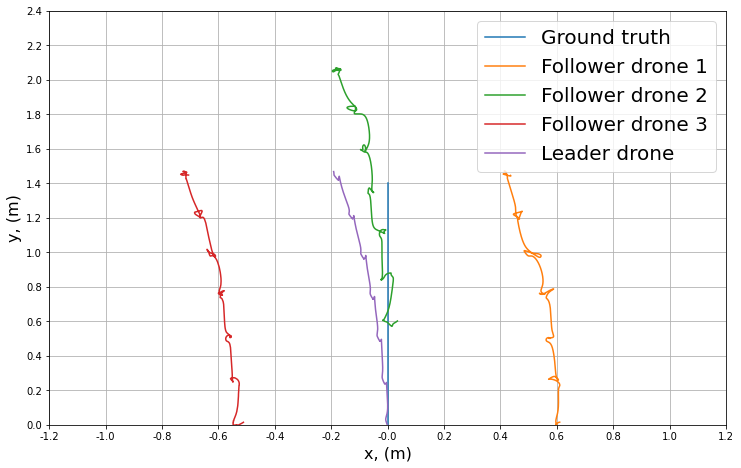}
 \caption{Linear trajectory following by a swarm, in which the leader drone is moving with Type 2 gait and $\beta = 45^{\circ}$.}
 \label{fig:results ape walk}
 \vspace{-0.4em}
\end{figure}

Table \ref{Table Result 2} shows the RMSE of 3.3 cm and maximal error does of 13.1 cm after one meter of linear path following. 

\begin{table}[h!]
\centering{
\caption{Positional error of the swarm. Gait Type 2. Star formation}
\setlength{\tabcolsep}{6pt} 
\renewcommand{\arraystretch}{1}
\begin{tabular}{ | c | c | c | }
\hline
\label{Table Result 2}
\begin{tabular}{p{1.5cm}cp{4cm}} \end{tabular} & {RMSE, m} & {Max Error, m} \\\hline
Leader Drone & 0.022 & 0.053 \\\hline
Follower drone 1 & 0.026 & 0.072 \\\hline
Follower drone 2 & 0.049 & 0.104 \\\hline
Follower drone 3 & 0.055 & 0.131 \\\hline
Followers overall & 0.043 & 0.131 \\\hline
Overall &0.033 & 0.131 \\\hline
\end{tabular}}
\end{table}

The experiments show the RMSE of the swarm formation in sufficient boundaries with Star topology and Type 2 gait. Type 1 gate showed a higher error of 0.039 m due to robot oscillation and velocity vector changing causing impedance link instability.

\section{Conclusion}
Using mathematical methods, we selected the parameters of the walking algorithm for acceptable stable robot motion, using data from the servos and data from tracking of the robot's motion along a given trajectory using the VICON system. We calculated the torques of the servomotors during using the structural dynamics equation for robot manipulators. During the experiments, 2 variants of robot walking were shown, subsequently the variant with the highest accuracy was chosen. The impedance control over the swarm of drones was investigated, and the Star topology was evaluated to ensure stable hand tracking by the drones (mean position error of 10.5 cm with 0.69 m/s maximal speed).Experimental results confirmed that the chassis could move steadily in a straight line with a standard deviation of length at each step of 195 mm and
$3^{\circ}$ standard yaw deviation. In the future we are planning to explore the navigation of the swarm taking into account it's dynamics and to develop a controller that eliminates deviations from the desired path.

\section{Acknowledgement}
The reported study was funded by RFBR and CNRS, project number 21-58-15006.
The authors are grateful to PhD student Sausar Karaph for his
support of the project.

\bibliographystyle{IEEEtran}
\bibliography{sample-base}
\end{document}